\documentclass{article}

\usepackage{PRIMEarxiv}
\usepackage[utf8]{inputenc}
\usepackage[T1]{fontenc}
\usepackage{natbib}
\usepackage{hyperref}
\usepackage{url}
\usepackage{booktabs}
\usepackage{amsmath}
\usepackage{amssymb}
\usepackage{amsfonts}
\usepackage{nicefrac}
\usepackage{microtype}
\usepackage{graphicx}
\usepackage{subcaption}
\usepackage{algorithm}
\usepackage{algpseudocode}
\usepackage{xcolor}
\usepackage{fancyhdr}

\graphicspath{{media/}}

\title{ThinkSwitch: Context Distillation with LoRA and Weight Interpolation for Specific-Purpose Reasoning Tasks}

\author{
  Dhruv Saini \\
  Bellevue High School \\
  \texttt{dhruv9saini@gmail.com}
  \And
  Rohan Pandey \\
  DigitalOcean \\
  \texttt{rpandey@digitalocean.com}
}

\begin{document}
\maketitle

\begin{abstract}
Large language models often improve on difficult tasks by spending inference-time compute on a reasoning trace before producing the final answer. That extra computation can be useful, but it also raises latency, token cost, and deployment complexity. We introduce \textbf{ThinkSwitch}, a low-compute procedure for co-training paired instruct and thinking checkpoints. Starting from compatible Qwen3-4B instruct and thinking models, each iteration asks the thinking checkpoint to generate answers, removes the reasoning trace, distills the answer-only pairs into the instruct checkpoint with QLoRA, and reconstructs a thinking checkpoint with spherical weight interpolation. The only human-supplied inputs are task prompts; the labels are generated by the model itself. On a 30-question AIME 2026 evaluation, ThinkSwitch improves the instruct checkpoint from 10/30 to 20/30 and the thinking checkpoint from 14/30 to 22/30. On a 30-question PubMedQA subset, it improves the instruct checkpoint from 13/30 to 18/30 and the thinking checkpoint from 18/30 to 25/30. The complete experiment uses 15 training prompts per domain and costs \$2.86 on a single cloud RTX 3070. The results are small-scale, but they indicate that targeted distillation loops can move part of the benefit of explicit reasoning into weights while preserving a separate thinking mode.
\end{abstract}

\keywords{Context Distillation \and Chain-of-Thought Reasoning \and Low-Rank Adaptation \and Model Merging \and Test-Time Compute}

\section{Introduction}

Reasoning models are expensive in a very literal sense: they spend tokens before they answer. A model that writes a long scratchpad can solve problems that a direct-answer model misses, but every extra token is paid for at inference time. In many cases the cost is part of the mechanism, since more computation is being allocated to the problem before the model commits to an answer~\citep{wei2022chain, kojima2022large, snell2024scaling}.

This creates a practical tension. For hard problems, we would like access to reasoning behavior without requiring every deployment path to emit a reasoning trace. We would also like post-training methods that fit within small-lab infrastructure rather than frontier-scale training clusters. Reinforcement learning can improve reasoning, as in DeepSeek-R1, but outcome rewards are sparse and rollouts are expensive at scale~\citep{deepseek2025r1}. Supervised fine-tuning gives a dense token-level signal, but it depends on data that is better than what the student can already produce~\citep{ouyang2022training, singh2025r2egym}. Iterated distillation and amplification offers another path toward self-improvement, although repeated amplification can be costly and broad-domain loops may plateau~\citep{christiano2018supervising}.

We focus on a narrower setting: a model family that provides two compatible checkpoints, one optimized for direct answers and one optimized for thinking. In that setting, can the thinking checkpoint serve as a teacher while the deployed direct-answer model remains cheap to run? ThinkSwitch addresses this with a four-step loop. The thinking checkpoint first generates solutions for a small prompt set; the reasoning traces are then removed; the direct-answer checkpoint is trained on the resulting answer-only pairs with QLoRA~\citep{hu2022lora, dettmers2023qlora}; and a new thinking checkpoint is formed by interpolating the updated instruct checkpoint with the previous thinking checkpoint using SLERP~\citep{shoemake1985animating, goddard2024mergekit}. The loop then repeats.

The central design choice is that the student never sees the teacher's scratchpad. This differs from rationale distillation, where the student is trained to reproduce intermediate reasoning. Our goal is not to make the instruct checkpoint write better reasoning traces. It is to make the instruct checkpoint answer more accurately when deployed in a direct-answer format, while preserving a separate thinking checkpoint for tasks where explicit deliberation remains useful.

The experiments are deliberately small. Each domain uses only 15 training prompts, with no human-written outputs, and the same loop is run for mathematics and biomedical question answering. Even under this constraint, performance improves on both AIME 2026 and PubMedQA. The main contributions are:
\begin{enumerate}
    \item We present ThinkSwitch, a simple loop that alternates trace-free QLoRA distillation with checkpoint interpolation between compatible instruct and thinking models.
    \item We show that answer-only distillation can improve both direct-answer and thinking checkpoints using only model-generated labels.
    \item We provide small-scale evidence on AIME 2026 and PubMedQA that the loop is data efficient, compute efficient, and sensitive to two necessary components: trace stripping and interpolation.
\end{enumerate}

\section{Background and Related Work}

\paragraph{Inference-time reasoning.}
Chain-of-thought prompting improves multi-step reasoning by making intermediate computation explicit~\citep{wei2022chain, kojima2022large}. Self-consistency further improves performance by sampling several reasoning paths and selecting an answer by agreement~\citep{wang2022self}. Recent work treats inference-time compute as a scaling axis in its own right~\citep{snell2024scaling}. The tradeoff is straightforward: longer traces cost more and add latency.

\paragraph{Context distillation.}
Context distillation trains a model to internalize behavior that was originally induced by extra context~\citep{askell2021general, snell2022learning}. Prior work shows that reasoning traces can sometimes be distilled into direct predictions~\citep{hsieh2023distilling, deng2024implicit, deng2024explicit}. ThinkSwitch follows this line, but uses paired thinking and instruct checkpoints and keeps the final system dual-mode rather than collapsing everything into one model.

\paragraph{Parameter-efficient fine-tuning.}
LoRA freezes a base weight matrix $W_0$ and learns a low-rank update $\Delta W = BA$, with $B \in \mathbb{R}^{d \times r}$ and $A \in \mathbb{R}^{r \times k}$ for $r \ll \min(d,k)$~\citep{hu2022lora}. The layer output is
\begin{equation}
    h = W_0 x + BAx .
\end{equation}
QLoRA reduces memory further by quantizing the frozen base model during adapter training~\citep{dettmers2023qlora}. These methods make repeated post-training loops feasible on commodity hardware.

\paragraph{Model interpolation and merging.}
Related checkpoints often occupy connected regions of weight space. Model soups average weights across fine-tuned models to improve accuracy without increasing inference cost~\citep{wortsman2022model}. Task arithmetic treats learned behaviors as approximately additive weight-space directions~\citep{ilharco2023editing, ortizjimenez2023task}. MergeKit operationalizes these ideas and supports spherical linear interpolation between checkpoints~\citep{goddard2024mergekit}. ThinkSwitch uses interpolation as the mechanism that carries newly distilled task signal back into a thinking checkpoint.

\paragraph{Position relative to self-improvement.}
ThinkSwitch is closest in spirit to iterative self-improvement methods such as STaR and iterated distillation~\citep{zelikman2022star, christiano2018supervising}. The difference is the unit of amplification. STaR-style methods amplify by sampling and training on rationales, whereas ThinkSwitch temporarily uses a thinking checkpoint and then compiles only the answer back into a direct-answer checkpoint. The procedure is therefore tied to model families that expose compatible instruct and thinking checkpoints, but for a narrow target distribution it is cheap enough to repeat.

\section{Method}

Let $A_0$ be the initial instruct checkpoint and $B_0$ be the initial thinking checkpoint. In our experiments, these are \texttt{Qwen3-4B-Instruct-2507} and \texttt{Qwen3-4B-Thinking-2507}~\citep{yang2025qwen3, qwen3instruct2507, qwen3thinking2507}. They share architecture and tokenizer. This compatibility is required: interpolation between unrelated models is not expected to preserve behavior.

At iteration $k$, ThinkSwitch constructs $A_{k+1}$ and $B_{k+1}$ from $A_k$ and $B_k$.

\paragraph{Teacher generation.}
For a prompt set $\mathcal{P}_k = \{x_i\}_{i=1}^{n}$, the thinking checkpoint generates a reasoning trace and final answer:
\begin{equation}
    (t_i, y_i) \sim p_{B_k}(\cdot \mid x_i).
\end{equation}
We decode deterministically with temperature 0 using vLLM~\citep{kwon2023efficient}. In this work, $n=15$ for each domain.

\paragraph{Trace stripping.}
Before training, the reasoning trace is removed. The distillation set is
\begin{equation}
    \mathcal{D}_k = \{(x_i, y_i)\}_{i=1}^{n}.
\end{equation}
For Qwen3 outputs, the implementation removes text enclosed by \texttt{<think>} and \texttt{</think>} tags. This step is central because the instruct model is not being asked to reproduce the scratchpad; it is trained only to map the prompt to the answer.

\paragraph{QLoRA distillation.}
We train LoRA adapters on $\mathcal{D}_k$ while keeping the base model quantized. If $\Delta_k$ denotes the adapter update learned at iteration $k$, the next instruct checkpoint is
\begin{equation}
    A_{k+1} = A_k + \Delta_k .
\end{equation}
This is supervised fine-tuning on model-generated, trace-free labels.

\paragraph{Thinking reconstruction.}
The next thinking checkpoint is formed by spherical interpolation between the updated instruct checkpoint and the previous thinking checkpoint. For flattened parameter vectors $u$ and $v$, SLERP with coefficient $\lambda$ is
\begin{equation}
    \operatorname{slerp}(u,v;\lambda)
    =
    \frac{\sin((1-\lambda)\Omega)}{\sin \Omega}u
    +
    \frac{\sin(\lambda\Omega)}{\sin \Omega}v ,
\end{equation}
where $\Omega = \arccos(u^\top v / \|u\|\|v\|)$. We set
\begin{equation}
    B_{k+1} = \operatorname{slerp}(A_{k+1}, B_k; \lambda).
\end{equation}
The role of this step is to keep the thinking checkpoint close to the original reasoning model while injecting the task-specific signal learned by the instruct checkpoint.

\subsection{Design Rationale}

\paragraph{Why answer-only distillation.}
The instruct checkpoint is ultimately evaluated in a direct-answer mode, so training it to reproduce \texttt{<think>} blocks would make the target format differ from the intended deployment behavior. Trace stripping makes the objective stricter: the model must map the prompt to the final answer without spending output tokens on a scratchpad. This does not prove that the model internally performs the same reasoning as the teacher, but it does test the operational property that matters for deployment: whether the answer improves when reasoning tokens are removed from the output channel.

\paragraph{Why keep two checkpoints.}
A single distilled checkpoint would force one compromise between speed and deliberation. ThinkSwitch instead maintains two modes: $A_k$ for short direct answers, and $B_k$ for cases where explicit reasoning or additional verification is worthwhile. This separation remains useful even if $A_k$ absorbs part of the reasoning benefit, since the main results show that the thinking checkpoint keeps a consistent advantage on the harder math setting.

\paragraph{Why interpolation after distillation.}
The QLoRA update changes the instruct checkpoint but does not by itself update the thinking checkpoint. Interpolation provides a cheap synchronization step between the two modes. The assumption is local: because $A_k$ and $B_k$ are homologous checkpoints from the same model family, a path between them in weight space can preserve useful behavior. We do not assume that arbitrary models can be interpolated this way; the compatibility requirement is a core limitation of the method.

\begin{algorithm}[t]
\caption{ThinkSwitch}
\label{alg:thinkswitch}
\begin{algorithmic}[1]
\Require Instruct checkpoint $A_0$, thinking checkpoint $B_0$, prompt batches $\{\mathcal{P}_k\}_{k=0}^{K-1}$
\For{$k = 0$ to $K-1$}
    \State Generate outputs $(t_i,y_i)$ from $B_k$ for each $x_i \in \mathcal{P}_k$
    \State Strip reasoning traces $t_i$ and form $\mathcal{D}_k = \{(x_i,y_i)\}$
    \State Train QLoRA adapter $\Delta_k$ on $\mathcal{D}_k$
    \State $A_{k+1} \gets A_k + \Delta_k$
    \State $B_{k+1} \gets \operatorname{slerp}(A_{k+1}, B_k; \lambda)$
\EndFor
\State \Return $A_K, B_K$
\end{algorithmic}
\end{algorithm}

\begin{figure}[t]
    \centering
    \includegraphics[width=0.88\linewidth]{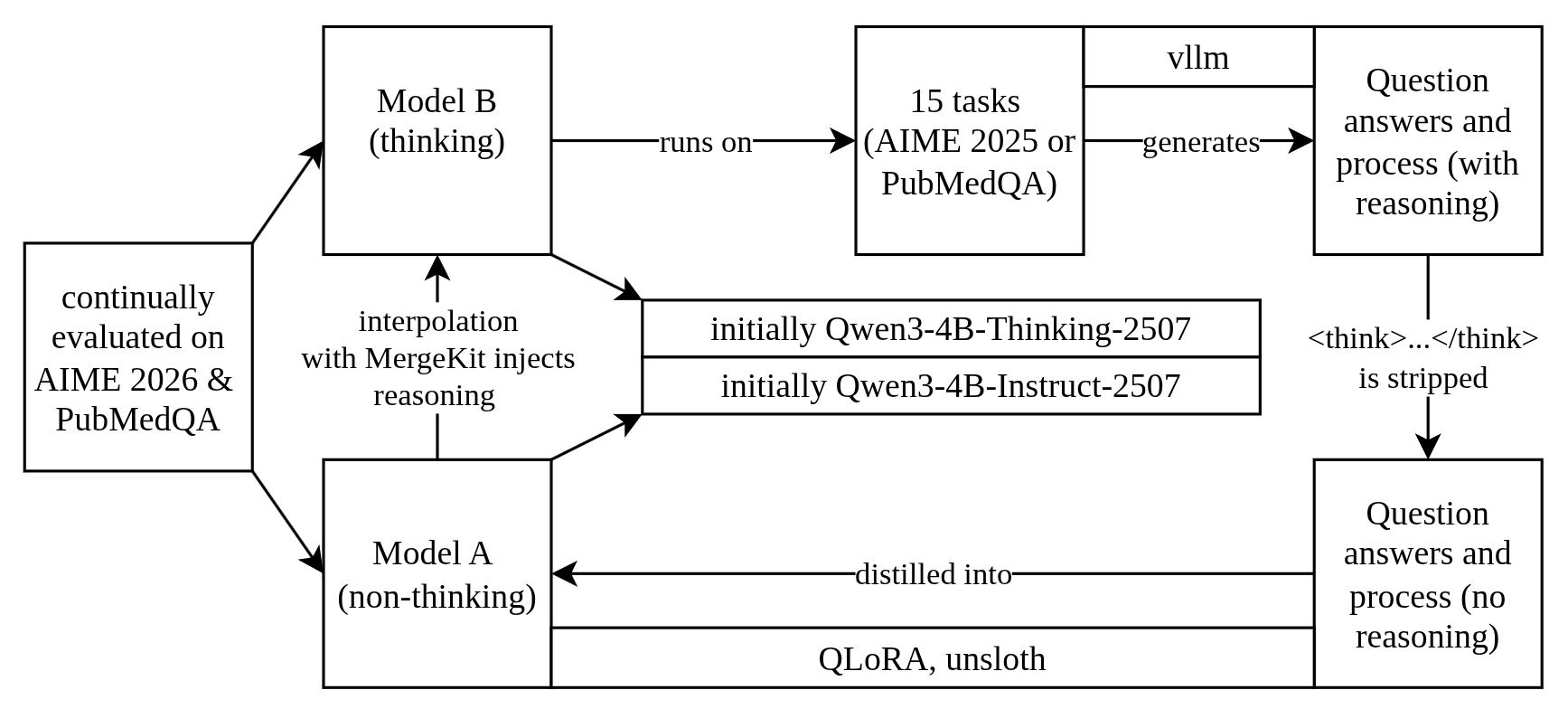}
    \caption{ThinkSwitch alternates teacher generation, trace stripping, QLoRA distillation, and thinking-checkpoint reconstruction.}
    \label{fig:flowchart}
\end{figure}

\section{Experimental Setup}

\paragraph{Models.}
All experiments use the Qwen3-4B 2507 instruct and thinking checkpoints~\citep{yang2025qwen3, qwen3instruct2507, qwen3thinking2507}. Qwen3 is appropriate for this study because it exposes closely related direct-answer and thinking variants.

\paragraph{Data.}
The mathematics run uses 15 training prompts drawn from AIME 2025 and evaluates on 30 AIME 2026 problems~\citep{aops2026aime}. The transfer run uses 15 PubMedQA training prompts and evaluates on a fixed 30-question PubMedQA subset~\citep{jin2019pubmedqa}. Only prompts are manually supplied. Training answers are generated by the current thinking checkpoint at each iteration. For PubMedQA, we first screen a 45-question candidate pool and select a 15-question training set on which the teacher answers all questions correctly. Appendix~\ref{app:figures} gives the dataset and screening details.

\paragraph{Training.}
Teacher outputs are generated with vLLM at temperature 0~\citep{kwon2023efficient}, and \texttt{<think>} blocks are removed before training. The instruct checkpoint is fine-tuned with QLoRA in Unsloth~\citep{dettmers2023qlora, han2025unsloth}. The training script uses rank $r=16$, LoRA alpha 16, dropout 0, batch size 1, gradient accumulation 8, bfloat16, and paged AdamW 8-bit. It targets the attention and MLP projection matrices. Each iteration runs for 3 epochs with learning rate $2 \times 10^{-4}$.

\paragraph{Evaluation.}
After each iteration, both $A_k$ and $B_k$ are evaluated. Instruct checkpoints answer directly. Thinking checkpoints may emit reasoning, but scoring uses only the final answer. We report exact-match correctness out of 30, with iteration 0 as the baseline.

\paragraph{Hardware and cost.}
All experiments run on a single cloud NVIDIA RTX 3070. The mathematics loop completes in under 24 hours. The total cost across both domains is \$2.86.

\section{Results}

\begin{figure}[t]
    \centering
    \begin{subfigure}{0.49\linewidth}
        \centering
        \includegraphics[width=\linewidth]{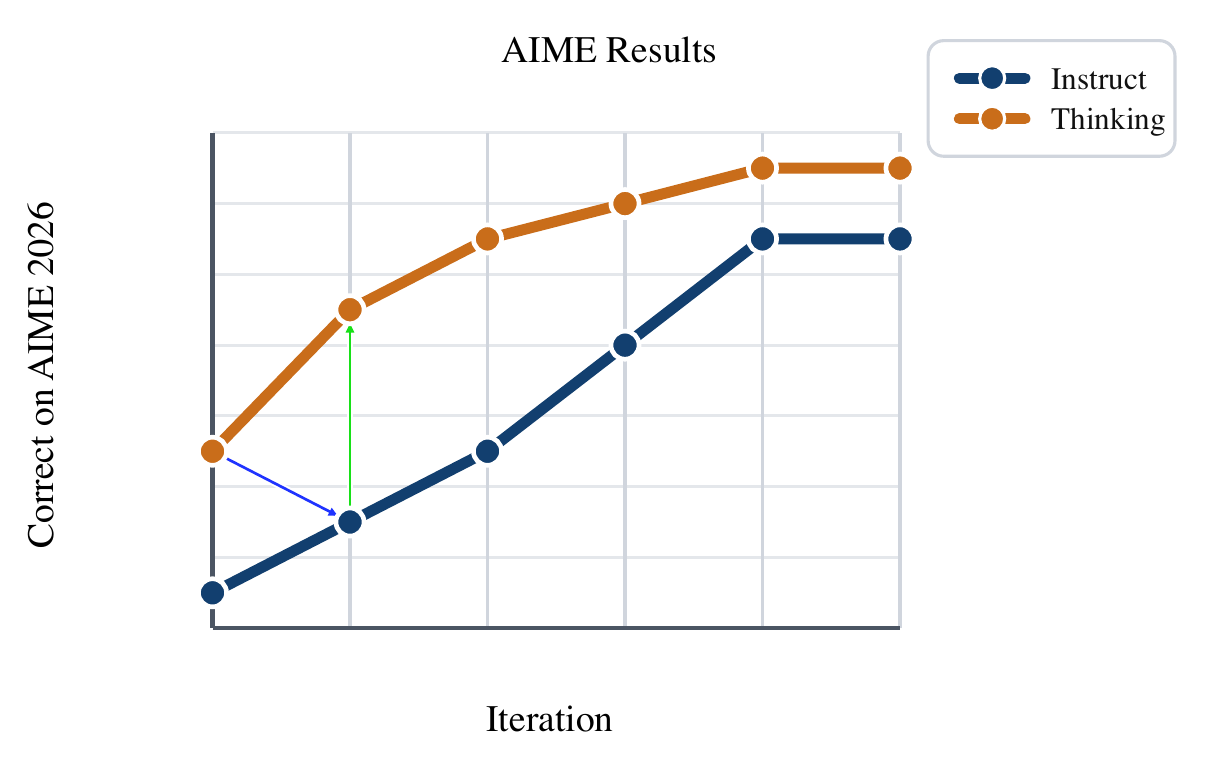}
        \caption{AIME 2026.}
    \end{subfigure}
    \hfill
    \begin{subfigure}{0.49\linewidth}
        \centering
        \includegraphics[width=\linewidth]{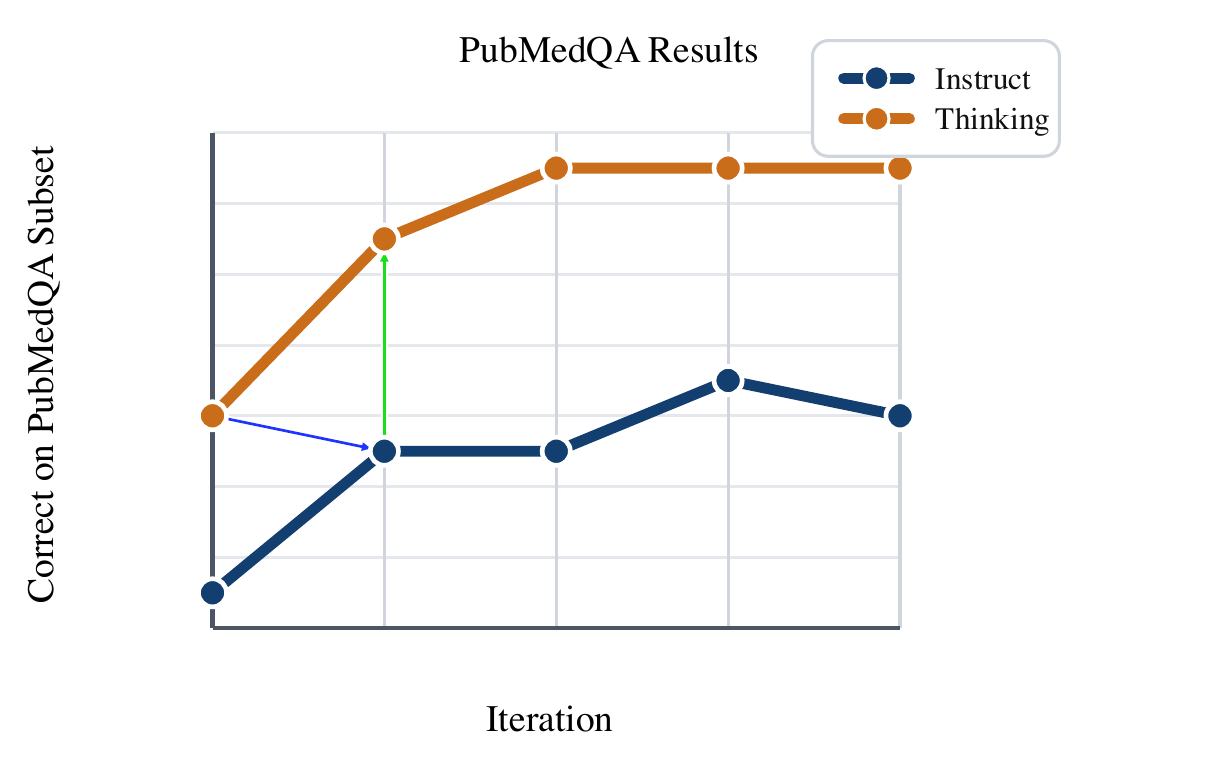}
        \caption{PubMedQA.}
    \end{subfigure}
    \caption{Accuracy trajectories across ThinkSwitch iterations for instruct and thinking checkpoints.}
    \label{fig:trajectories}
\end{figure}

Table~\ref{tab:main-results} reports baseline, best, final, and net gains for both domains. The same runs are plotted over all iterations in Figure~\ref{fig:trajectories}.

\begin{table}[t]
\centering
\caption{Baseline, best, and final scores. Each evaluation has 30 questions.}
\label{tab:main-results}
\begin{tabular}{lcccc}
\toprule
Run & Baseline & Best & Final & Net gain \\
\midrule
AIME 2026 instruct & 10/30 & 20/30 & 20/30 & +10 \\
AIME 2026 thinking & 14/30 & 22/30 & 22/30 & +8 \\
PubMedQA instruct & 13/30 & 19/30 & 18/30 & +5 \\
PubMedQA thinking & 18/30 & 25/30 & 25/30 & +7 \\
\bottomrule
\end{tabular}
\end{table}

\paragraph{AIME 2026.}
On AIME 2026, the instruct checkpoint rises from 10/30 to 20/30, while the thinking checkpoint rises from 14/30 to 22/30. Both improve through the early iterations and then flatten. That plateau is useful evidence because it suggests that the loop saturates on the available 15-prompt signal rather than becoming unstable. The gap between thinking and instruct narrows from 4 points at baseline to 2 points at convergence, consistent with answer-only distillation transferring part, but not all, of the reasoning advantage into direct-answer behavior.

\paragraph{PubMedQA.}
On PubMedQA, the thinking checkpoint rises from 18/30 to 25/30 and keeps that score through the final iteration. The instruct checkpoint peaks at 19/30 and ends at 18/30. This asymmetry is consistent with mild overfitting in the direct-answer checkpoint: the thinking checkpoint remains anchored by interpolation with the reasoning model, while the instruct checkpoint receives repeated updates from a very small answer-only set.

\paragraph{Statistical analysis.}
We compare baseline and final scores using one-sided Fisher exact tests and combine instruct and thinking p-values with Fisher's method. The combined p-value is 0.0028 for AIME 2026 and 0.0384 for PubMedQA. Because each evaluation has only 30 questions, these values should be read as evidence for the observed runs rather than as a complete benchmark claim. Still, the gains are large relative to the evaluation size.

\paragraph{Efficiency.}
The full experiment uses 15 prompts per domain, one cloud RTX 3070, and \$2.86 of total compute. Training loss curves and a cost-frontier visualization are provided in Appendix~\ref{app:figures}. These supporting plots are not substitutes for held-out evaluation, but they show that the method fits within the resource profile of small-lab or individual research.

\section{Ablations}

We evaluate two ablations. In the first, trace stripping is removed and the model is trained on outputs that still contain reasoning traces. Both checkpoints regress immediately and do not recover. This indicates that the answer-only format is not cosmetic: it matches the deployment behavior expected from an instruct model and prevents the student from spending capacity imitating scratchpad style.

In the second ablation, interpolation is removed. The instruct checkpoint improves slightly for one iteration, but the overall loop collapses toward near-zero scores. This suggests that interpolation is the mechanism that keeps the two checkpoints aligned. Without it, the instruct checkpoint moves under QLoRA updates while the thinking checkpoint no longer receives the distilled signal in a compatible form.

These ablations should be interpreted as stress tests rather than exhaustive component studies. They do not identify the optimal interpolation coefficient, prompt-selection policy, or LoRA configuration. They do, however, show that the two most unusual parts of the method are not optional in this setting: removing traces protects the instruct format, and interpolation keeps the two modes synchronized.

\section{Discussion}

ThinkSwitch is not a replacement for large-scale reinforcement learning. It is a small-loop method for settings where a compatible instruct and thinking pair already exists. The method is strongest when the task distribution is narrow, teacher answers are usually correct, and the output can be scored or extracted reliably. Within that regime, the main advantage is iteration speed: a researcher can run the loop, inspect failures, adjust the prompt pool, and rerun without building a reward environment or collecting human-written solutions.

The results also show a limitation: 15 prompts are enough to produce measurable gains, but not enough to avoid saturation. Once the prompt pool is exhausted, later iterations reinforce the same patterns. This behavior is visible in AIME, where both modes flatten after early gains. A natural extension is curriculum selection, either by adding harder prompts when validation gains flatten or by replacing prompts that no longer produce useful teacher signal. In this view, ThinkSwitch is less a one-shot training recipe than a mechanism for converting a small stream of targeted prompts into paired checkpoint improvements.

The PubMedQA run suggests that the method is not specific to short numerical answers. The biomedical setting has different surface form, answer distribution, and prior knowledge requirements. At the same time, PubMedQA remains a constrained question-answering benchmark. Broader validation is needed on coding, science, and open-domain tasks where answer extraction is harder and partial correctness matters.

\paragraph{Limitations.}
The evaluation sets are small, with 30 questions per domain, so the results should be viewed as a focused proof of concept rather than a leaderboard claim. The method also depends on compatible paired checkpoints. If a model exposes thinking behavior only through prompting or an inference-time flag, there may be no separate checkpoint to interpolate. Finally, answer extraction is heuristic: the current trace stripper removes \texttt{<think>} blocks and keeps the remaining answer. Malformed generations can introduce noisy labels, which is more consequential when the training set contains only 15 examples.

\paragraph{Future work.}
Future work should test larger prompt pools, stronger model pairs, multiple interpolation coefficients, and additional domains. The most important next step is to replace fixed prompt sets with adaptive curricula. A second direction is structured output parsing, so that teacher answers can be extracted and verified before they enter the distillation set. A third direction is measuring mode drift: after many rounds of QLoRA and interpolation, the instruct and thinking checkpoints may eventually move far enough apart that interpolation no longer preserves coherent behavior.

\section{Conclusion}

ThinkSwitch is a low-compute method for improving paired instruct and thinking checkpoints through trace-free QLoRA distillation and spherical weight interpolation. Using only 15 prompts per domain, it doubles AIME 2026 instruct accuracy from 10/30 to 20/30, improves AIME thinking accuracy from 14/30 to 22/30, and transfers to PubMedQA with gains for both instruct and thinking modes. The experiment costs \$2.86 on a single RTX 3070. Taken as a focused proof of concept, these results suggest that targeted, trace-free self-distillation can move useful reasoning behavior into weights while keeping a separate thinking mode available.

\section*{Acknowledgments}
We thank the Summer Research Opportunity Program (SROP) for supporting this collaboration.

\bibliographystyle{unsrtnat}
\bibliography{references}

\appendix
\section{Additional Experimental Figures}
\label{app:figures}

\begin{figure}[h]
    \centering
    \begin{subfigure}{0.49\linewidth}
        \centering
        \includegraphics[width=\linewidth]{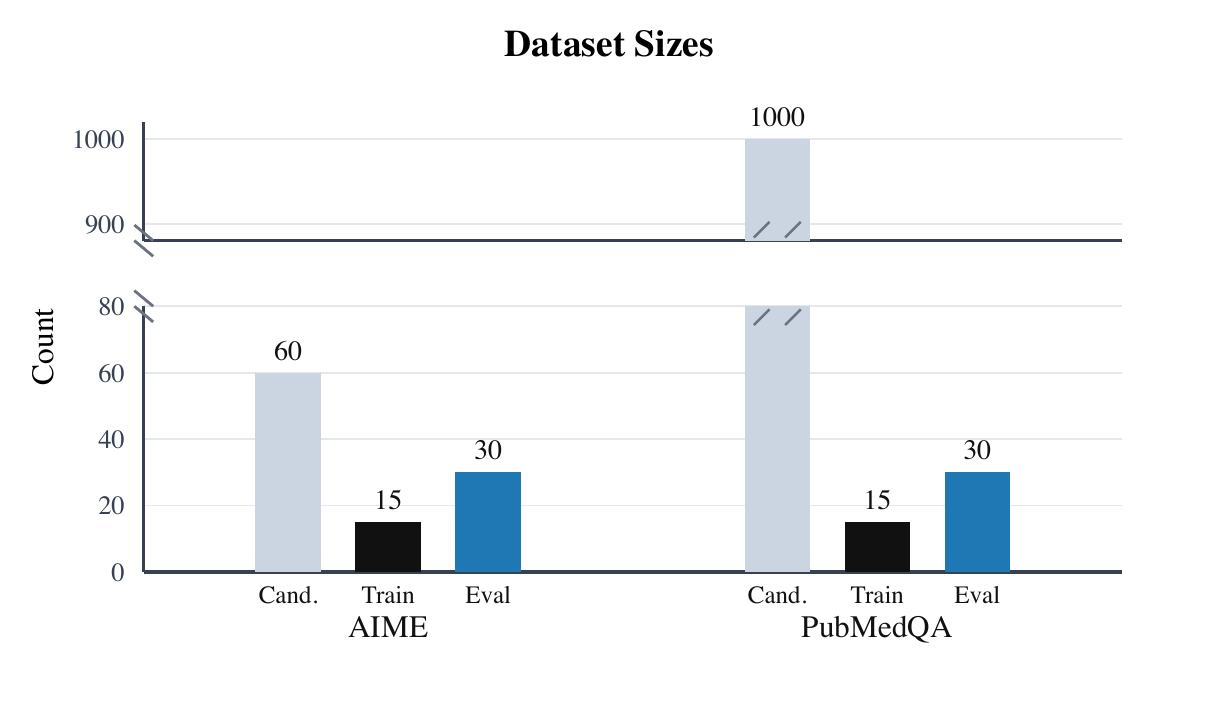}
        \caption{Dataset sizes.}
    \end{subfigure}
    \hfill
    \begin{subfigure}{0.49\linewidth}
        \centering
        \includegraphics[width=\linewidth]{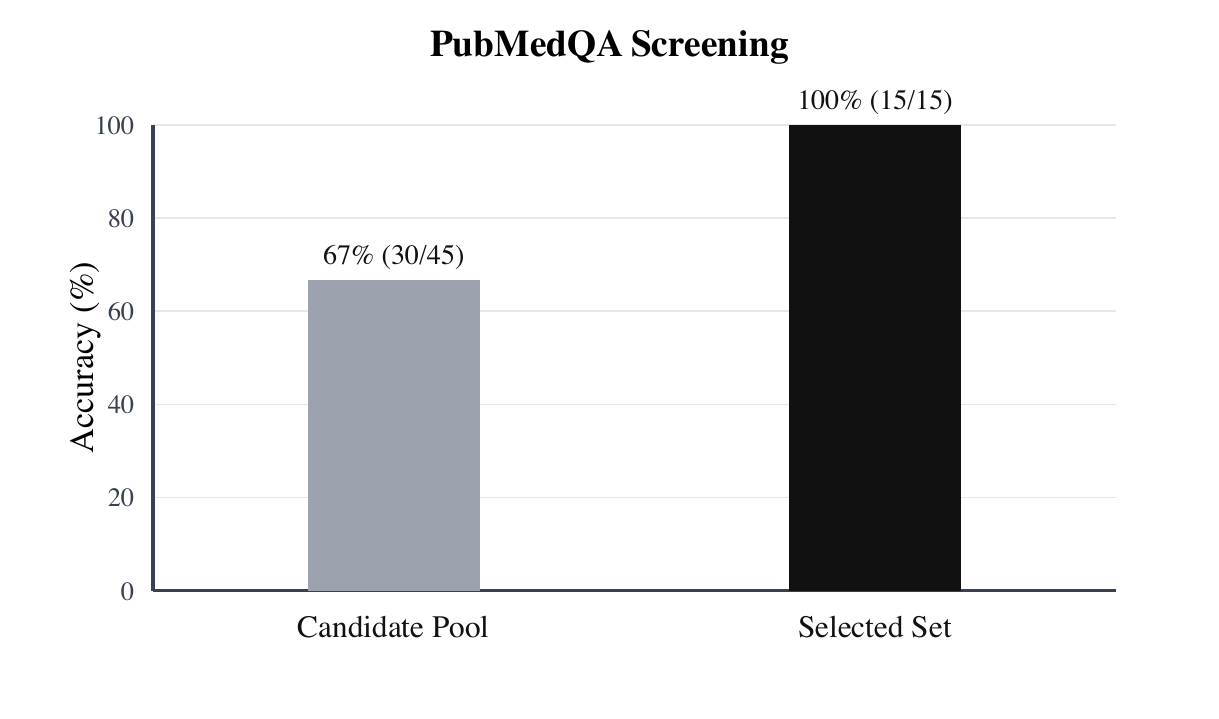}
        \caption{PubMedQA screening.}
    \end{subfigure}
    \caption{Training uses 15 prompts per domain and evaluates on 30 held-out questions per domain. For PubMedQA, the selected training set is drawn from a screened candidate pool.}
    \label{fig:data}
\end{figure}

\begin{figure}[h]
    \centering
    \begin{subfigure}{0.49\linewidth}
        \centering
        \includegraphics[width=\linewidth]{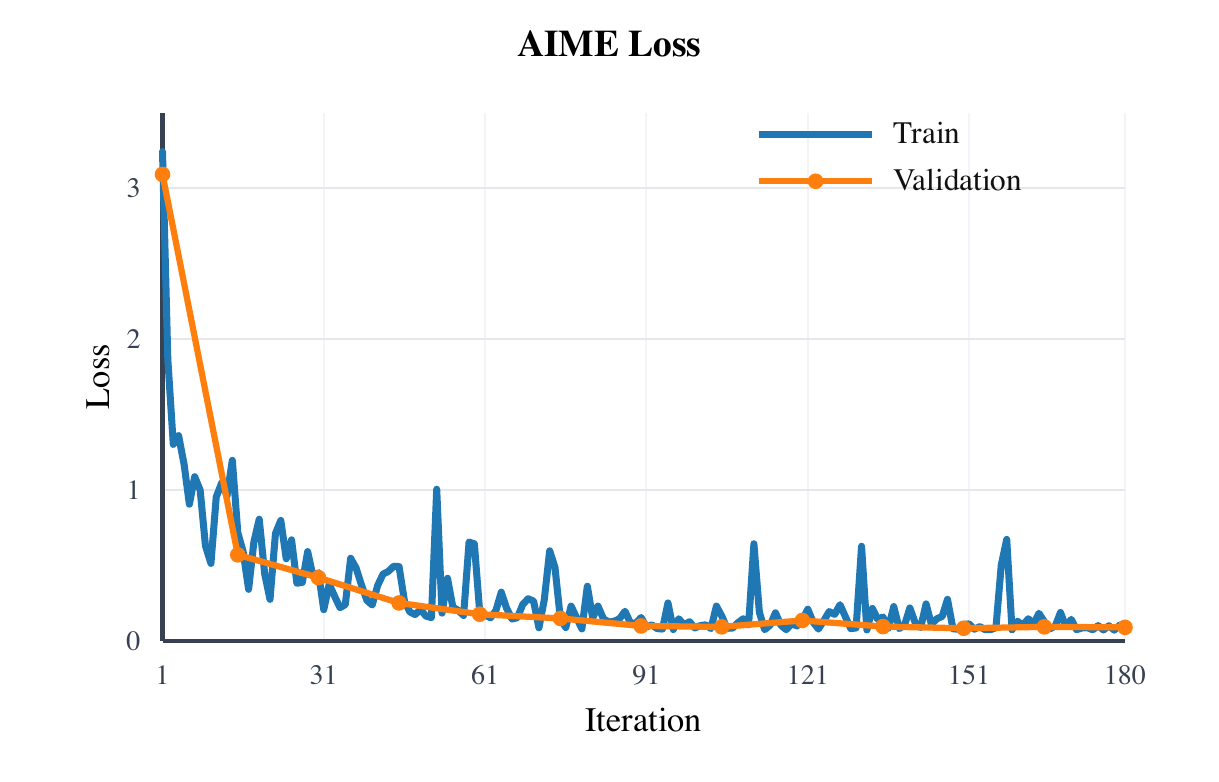}
        \caption{AIME training.}
    \end{subfigure}
    \hfill
    \begin{subfigure}{0.49\linewidth}
        \centering
        \includegraphics[width=\linewidth]{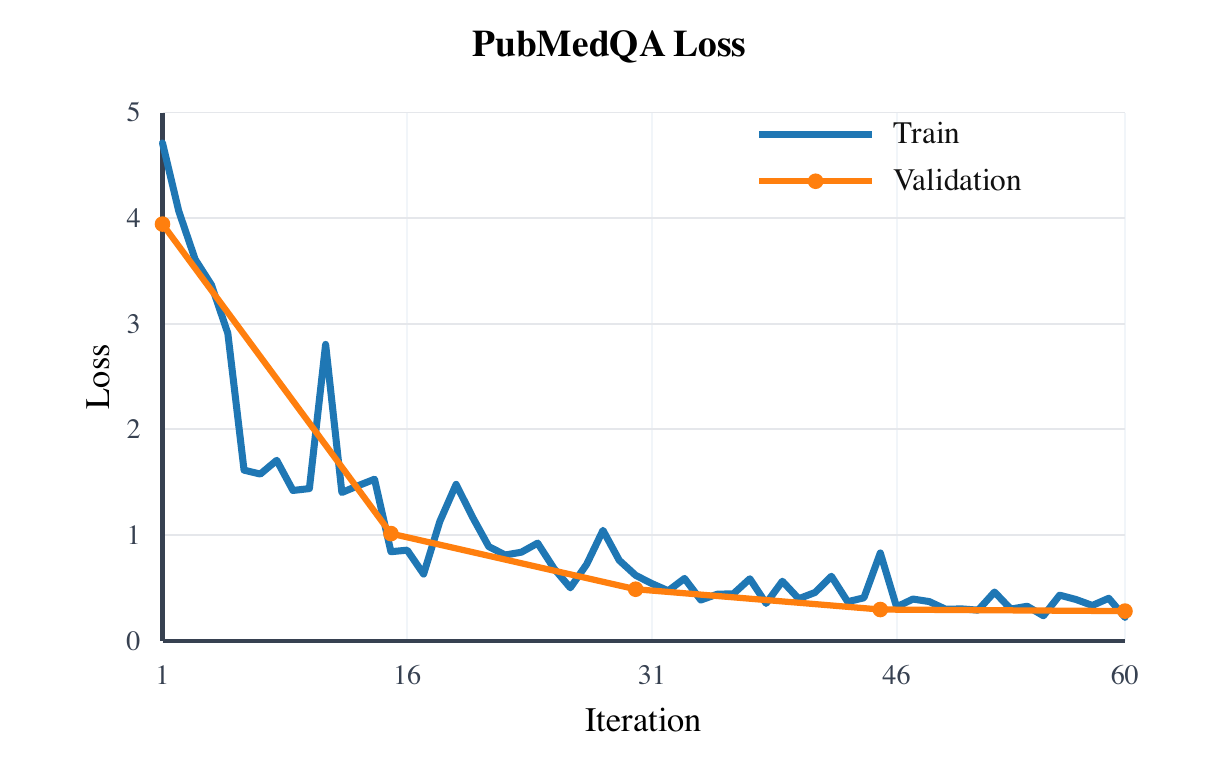}
        \caption{PubMedQA training.}
    \end{subfigure}
    \caption{QLoRA training and validation losses for the trace-free answer-only datasets. The curves show that the small datasets are fit quickly; held-out exact-match evaluation is therefore the primary evidence of improvement.}
    \label{fig:loss}
\end{figure}

\begin{figure}[h]
    \centering
    \includegraphics[width=0.82\linewidth]{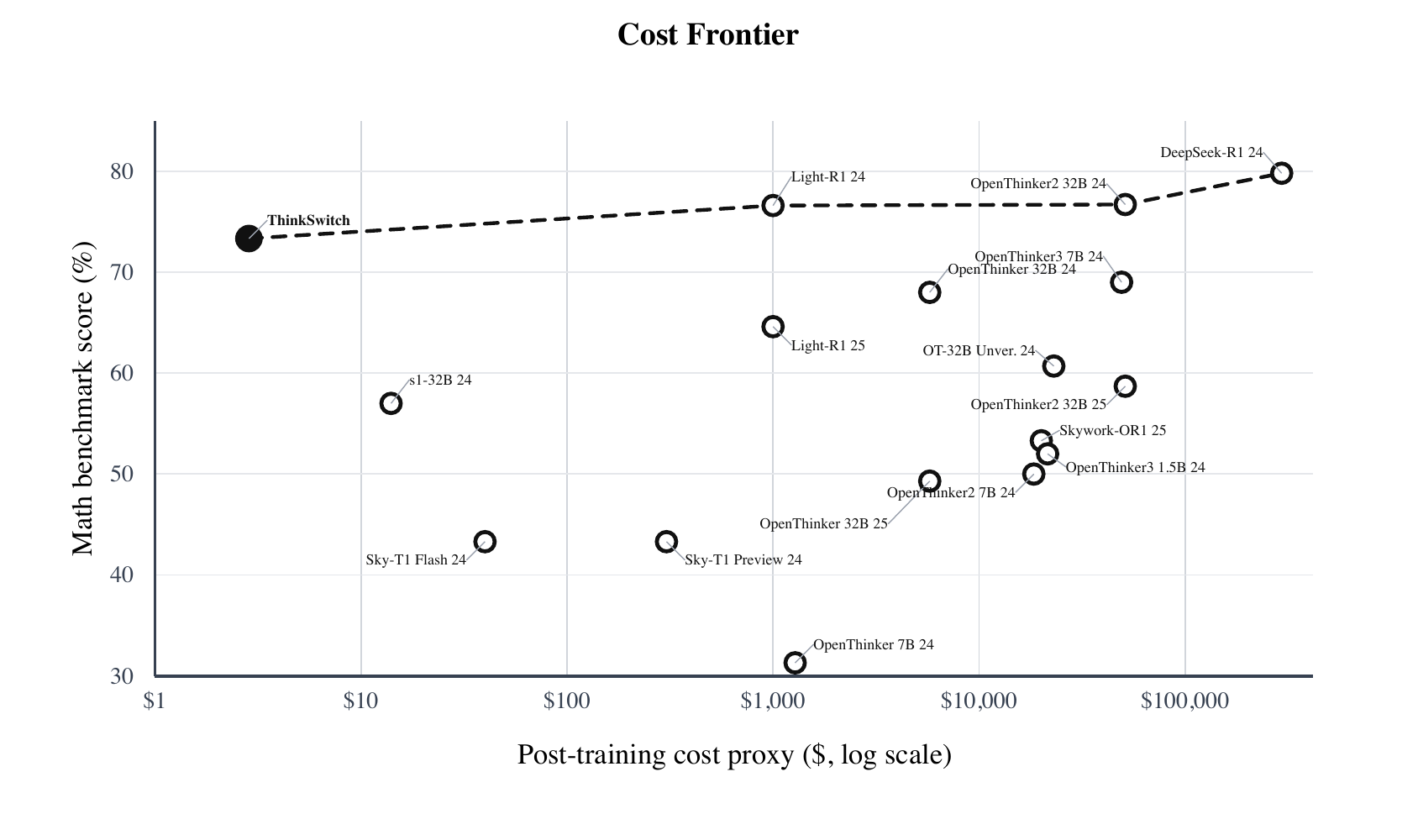}
    \caption{Post-training cost proxy against math benchmark score. The comparison is approximate because public reports use different cost accounting methods, but it places ThinkSwitch in the low-cost region of the small-model post-training frontier.}
    \label{fig:cost}
\end{figure}

\end{document}